\documentclass{article}
\usepackage{ijcai17}

\usepackage{times}

\usepackage{amssymb}
\usepackage{amsmath}
\usepackage{gensymb}
\usepackage{algorithm}
\usepackage{algorithmic}
\usepackage{graphicx}
\title{Improving Deep Learning using Generic Data Augmentation}
\author{Luke Taylor\\ 
Department of Computer Science, University of Cape Town\\
pcchair@ijcai-17.org}
\author{Luke Taylor \\ University of Cape Town \\ Department of Computer Science \\ Cape Town, South Africa \\ tylchr011@uct.ac.za \And
               Geoff Nitschke \\ University of Cape Town \\ Department of Computer Science \\ Cape Town, South Africa \\ gnitschke@cs.uct.ac.za}

\begin{document}

\maketitle

\begin{abstract}
Deep artificial neural networks require a large corpus of training data in order to
effectively learn, where collection of such training data is often expensive
and laborious.
\textit{Data augmentation} overcomes this issue by artificially inflating
the training set with label preserving transformations.
Recently there has been extensive use of generic data augmentation to improve
\textit{Convolutional Neural Network} (CNN) task performance.
This study benchmarks various popular data augmentation schemes to allow researchers
to make informed decisions as to which training methods are most appropriate for
their data sets.  Various geometric and photometric
schemes are evaluated on a coarse-grained data set using a relatively simple
CNN.  Experimental results, run using 4-fold cross-validation and reported in
terms of Top-1 and Top-5 accuracy, indicate that \textit{cropping} in
\textit{geometric augmentation} significantly increases CNN task performance.
\end{abstract}

\section{Introduction}

\textit{Convolutional Neural Networks} (CNNs) are \cite{Handwritten}
synonymous with deep learning, a hierarchical model of learning with multiple levels of representations,
where higher levels capture more abstract concepts.
A CNNs connectivity pattern, inspired by the animal visual cortex,
enables the recognition and learning of spatial data such as images \cite{Handwritten},
audio \cite{SpeechRecognition} and text \cite{Text}.
With recent developments of large data sets and increased computing power,
CNNs have managed to achieve state-of-the-art results in various computer vision tasks
including large scale image and video classification \cite{DeepFace}.
However, an issue is that most large data sets are not publicly available and training a CNN on small data-sets
makes it prone to over-fitting, inhibiting the CNNs capability to generalize to unseen invariant data.
%\indent

A potential solution is to make use of \textit{Data Augmentation} (DA) \cite{Yaeger}, which is a regularization scheme that artificially
inflates the data-set by using label preserving transformations to add more invariant examples.
Generic DA is a set of computationally inexpensive methods \cite{AlexNet}, previously used to reduce over-fitting in training a CNN for the
\textit{ImageNet Large-Scale Visual Recognition Challenge} (ILSVRC) \cite{ImageNet}, and achieved state-of the-art results at the time.
This augmentation scheme consists of \textit{Geometric} and \textit{Photometric} transformations \cite{KaimingSun2014},
\cite{SimonyanZisserman2014}.

Geometric transformations alter the geometry of the image with the aim of making the CNN invariant to change in position and orientation.
Example transformations include flipping, cropping, scaling and rotating.
Photometric transformations amend the color channels with the objective of making the CNN invariant to change in lighting and color.
For example, color jittering and \textit{Fancy Principle Component Analysis (PCA)} \cite{AlexNet}, \cite{Howard}.
%which performs PCA on the set of RGB pixel values
%throughout the training set and adding multiples of the found principal components to the training images which resulted in a reduction of over $1\%$ in the top-1 error rate.
%The second form of augmentation consisted of a photometric transformation which has become to be known as \textit{Fancy PCA}.
%This scheme consists of performing Principle Component Analysis (In the next year \cite{Howard} improved upon these results by using DA which consisted of a more robust cropping scheme aided by extensive color manipulation.\\

Complex DA is a scheme that artificially inflate the data set by using domain specific synthesization to produce more training data.
This scheme has become increasingly popular \cite{Synthesization1,Synthesization2}
as it has the ability to generate \textit{richer} training data compared to the generic augmentation methods.
For example, Masi et al. \cite{FaceSynthesization} developed a facial recognition system using synthesized faces with different poses and
facial expressions to increase appearance variability enabling comparable task performance to state of the art facial recognition systems
using less training data.
At the frontier of data synthesis are \textit{Generative Adversarial Networks} (GANs) \cite{GAN} that have the ability to generate new samples after
being trained on samples drawn from some distribution.
For example, Zhang et al. \cite{GANExample} used a stack construction of GANs to generate realistic images of birds and flowers from text descriptions.

% to the dataset and can be categorised as follows:
%CNNs where developed by  for the task of recognising handwritten digits.

\begin{figure}[t]
    \centering
    \includegraphics[width=0.5\textwidth]{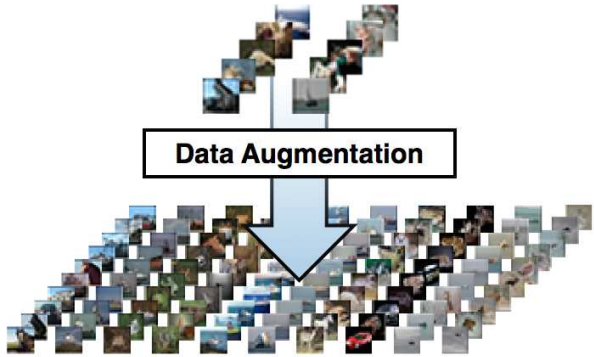}
    \caption{\textit{Data Augmentation} (DA) artificially inflates data-sets using label preserving transformations.}
    \label{fig:intro}
\end{figure}

%\begin{enumerate}
%\item ...
%\begin{enumerate}
%	\item Geometric Transformations: These are transformations that alter the geometry of the image with the aim of making the CNN invariant to change in position and orientation. Example transformations include flipping, cropping, scaling and rotating.
%	\item Photometric Transformations: These are transformations that amend the color channels with the objective of making the CNN invariant to change in lighting and color. Examples include color jittering and fancy PCA.
%	\end{enumerate}
%\item Complex DA: These are schemes that artificially inflate the data set by using domain specific synthesization to produce more training data. This has become an increasingly popular field of study \cite{Synthesization1,Synthesization2} as it has the ability to generate \textit{richer} training data compared to the generic augmentation methods. An example is \cite{FaceSynthesization} that developed a facial recognition system which uses synthesised faces with different poses and facial expressions to increase appearance variability enabling them to match state of the art facial recognition systems using less training data. At the frontier of data synthesis are Generative Adversarial Networks (GANs) that have the ability to generate new samples after being trained on samples drawn from some distribution \cite{GAN}. Notable is \cite{GANExample} which use a stack construction of GANs to generate realistic images of birds and flowers from text descriptions.
%\end{enumerate}

Thus, DA is a scheme to further boost CNN performance and prevent over-fitting.
The use of DA is especially well-suited when the training data is limited or laborious to collect.
Complex DA, albeit being a powerful augmentation scheme, is computationally expensive and time-consuming to implement.
A viable option is to apply generic DA as it is computationally inexpensive and easy to implement.

Prevalent studies that comparatively evaluate various popular DA methods include those outlined in table
\ref{table:one} and described in the following.
Chatfield et al. \cite{ReturnDevil} addressed how different CNN architectures compared to each other by
evaluating them on a common data-set.
This study was mainly focused on rigorous evaluation of deep architectures and shallow encoding methods, though an
evaluation of three augmentation methods was included.
These consisted of flipping and combining flipping and cropping on training images
in the coarse grained \textit{Caltech 101}\footnote{$www.vision.caltech.edu/Image Datasets/Caltech101/$}
and \textit{Pascal VOC}\footnote{$host.robots.ox.ac.uk/pascal/VOC/$} data-sets.
In additional experiments the authors trained a CNN using gray-scale images, though lower task performance was observed.
Overall results indicated that combining flipping and cropping yielded an increased task performance of $2 \sim 3\%$.
A shortcoming of this study was the few DA methods evaluated.
% and the lack of color information in the data-sets.
%DA is a powerful scheme to improve CNN performance, however no benchmark of  has been conducted.
%Some studies have started to investigate the effect of DA yet miss certain coverage: \textbf{(1)}
%They reported that flipping improves performance by a marginal $1\%$ however the combination of flipping and cropping results in performance of $2 \sim 3\%$. They noted that a higher density of crops from the centre of the image does not relate to improved performance.
%Additionally they performed training on just grayscale values of the images and noted a decrease of $3\%$ which indicates the importance of the color information. The shortcoming of this study in the context of DA benchmarking was the limited amount of DA methods explored.

Mash et al. \cite{AircraftDA} bench-marked a variety of geometric augmentation methods for the task of aircraft classification,
using a fine-grained data-set of $10$ classes.
Augmentation methods tested included cropping, rotating, re-scaling,
polygon occlusion and combinations of these methods.
The cropping scheme combined with occlusion yielded the most benefits, achieving a $9\%$ improvement over a benchmark
task performance.
Although this study evaluated various DA methods, photometric methods were not investigated and none were bench-marked on a coarse-grained
data-set.

In line with this work, further research \cite{Galaxy} noted that certain augmentation methods benefit from fine-grained data-sets.
For example, extensive use of rotating training images to increase CNN task performance for galaxy morphology classification
using a fine-grained data-set \cite{Galaxy}.

%It has to be noted that benefit more from An example is  where they make
%In order to get a better idea of the effectiveness of certain DA methods and how they generalize to other datasets they should be benchmarked on a coarse-grained dataset rather than a fine-grained one. This paper wishes fill the gap in current literature by addressing the need of benchmarking a variety of popular DA methods on a coarse-grained dataset.

However, to date, there has been no comprehensive studies that comparatively evaluate various popular DA methods on large
coarse-grained data-sets in order to ascertain the most appropriate DA method for any given data-set.
Hence, this study's objective is to evaluate a variety of popular geometric and photometric augmentation schemes on the coarse grained
Caltech101 data-set.
Using a relatively simple CNN based on that used by Zeiler and Fergus \cite{VisualizingCNN}, the goal is to contribute to empirical
data in the field of deep-learning to enable researchers to select the most appropriate generic augmentation scheme for a given data-set.

\begin{table}[t]
\centering
 \begin{tabular}{ c c c c }
 \hline
 & \textbf{Coarse-} & \textbf{Geometric} & \textbf{Photometric} \\ [0.5ex]
 & \textbf{Grained} & \textbf{DA}        & \textbf{DA} \\
 \hline\hline
 Chatfield \textit{et al.} & \checkmark & \checkmark & \\
 Mash \textit{et al.}      &  & \checkmark & \\
 \textbf{Our benchmark} & \checkmark & \checkmark & \checkmark \\ [0.5ex]
 \hline
 \end{tabular}
 \caption{Properties of studies that investigate DA.}
\label{table:one}
\end{table}

\newpage

\section{Data Augmentation (DA) Methods}
\begin{figure*}[t]
    \centering
    \includegraphics[width=0.90\textwidth]{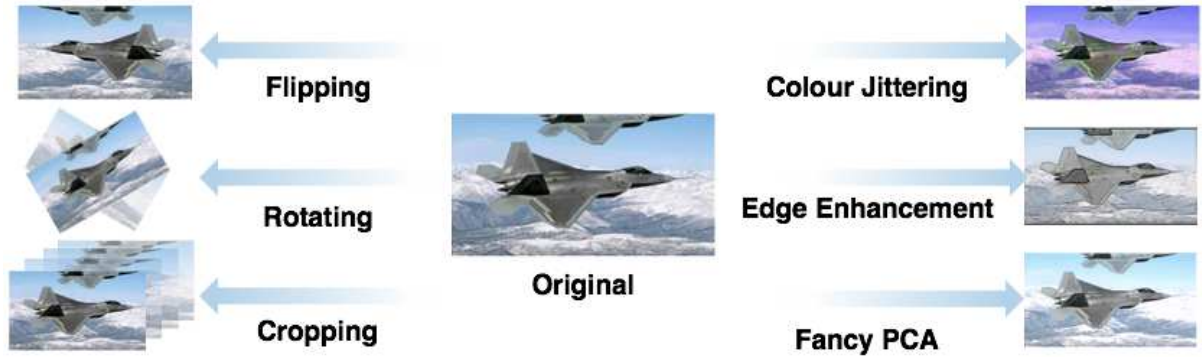}
    \caption{Overview of the \textit{Data Augmentation} (DA) methods evaluated.}
    \label{fig:overview}
\end{figure*}

DA refers to any method that artificially inflates the original training set with label preserving transformations and can be represented as the
mapping:
\[ \phi: \mathcal{S} \mapsto \mathcal{T} \]
Where, $\mathcal{S}$ is the original training set and $\mathcal{T}$ is the augmented set of $\mathcal{S}$.
The artificially inflated training set is thus represented as:
\[ \mathcal{S'} = \mathcal{S} \cup \mathcal{T} \]
Where, $\mathcal{S'}$ contains the original training set and the respective transformations defined by $\phi$.
Note the term label preserving transformations refers to the fact that if image $x$ is an element of class $y$ then $\phi(x)$ is also an element of class $y$.\\
\indent
As there is an endless array of mappings $\phi(x)$ that satisfy the constraint of being label preserving,
this paper evaluates popular augmentation methods used in recent literature \cite{ReturnDevil}, \cite{AircraftDA}
as well as a new augmentation method (section \ref{sec:photoMethods}).
Specifically, seven augmentation methods were evaluated (figure \ref{fig:overview}),
where one was defined as being \textit{No-Augmentation} which acted as the task performance benchmark for all the experiments given
three geometric and three photometric methods.

\subsection{Geometric Methods}
These are transformations that alter the geometry of the image by mapping the individual pixel values to new destinations. The  underlying shape of the class represented within the image is preserved but altered to some new position and orientation.
Given their success in related work \cite{AlexNet,Galaxy,Rotate2} we investigated the \textit{flipping}, \textit{rotation}
and \textit{cropping} schemes.\\
\indent
Flipping mirrors the image across its vertical axis. It is one of the most used augmentation schemes after being popularized
by Krizhevsky et al. \cite{AlexNet}.
It is computationally efficient and easy to implement due to only requiring rows of image matrices to be reversed.\\
\indent
The rotation scheme rotates the image around its center via mapping each pixel $(x,y)$ of an image to $(x',y')$ with
the following transformation \cite{Galaxy,Rotate2}:

\[ \begin{pmatrix}
{x}'\\ {y}'
\end{pmatrix}
=\begin{pmatrix}
cos\theta & -sin\theta\\
sin\theta & cos\theta
\end{pmatrix}
\begin{pmatrix}
x\\ y
\end{pmatrix} \]

Where, exploratory experiments indicated that setting $\theta$ as $-30\degree$ and $+30\degree$ establishes rotations that are large enough to generate new invariant samples. \\
\indent
Cropping is another augmentation scheme popularized by Krizhevsky et al. \cite{AlexNet}.
We used the same procedure as in related work \cite{ReturnDevil},
which consisted of extracting $224\times224$ crops from the four corners and the center of the $256\times256$ image.

\subsection{Photometric Methods}\label{sec:photoMethods}

These are transformations that alter the $RGB$ channels by shifting each pixel value $(r,g,b)$ to new pixel values $(r',g',b')$
according to pre-defined heuristics.
This adjusts image lighting and color and leaves the geometry unchanged.
We investigated the \textit{color jittering}, \textit{edge enhancement} and \textit{fancy PCA} photometric methods.\\
\indent
Color jittering is a method that either uses random color manipulation \cite{Jitter} or set color adjustment \cite{Rotate2}.
We selected the latter due to its accessible implementation using a HSB filter\footnote{www.jhlabs.com/ip/filters/HSBAdjustFilter.html}.\\
\indent
Edge enhancement is a new augmentation scheme that enhances the contours of the class represented within the image.
As the learned kernels in the CNN identify shapes it was hypothesized that CNN performance could be boosted by providing training images with intensified contours.
This augmentation scheme was implemented as presented in \textit{Algorithm 1}.
Edge filtering was accomplished using the \textit{Sobel} edge detector \cite{GonzalezWoods1992}
where edges were identified by computing the local gradient $\nabla \mathcal{S}(i,j)$ at each pixel in the image $\mathcal{S}$.\\

%\begin{algorithm}
\begin{algorithmic}
\small
\REQUIRE Source image: $\mathcal{I}$
\STATE $\mathcal{T} \leftarrow$ edge filter $\mathcal{I}$
\STATE $\mathcal{T} \leftarrow$ grayscale $\mathcal{T}$
\STATE $\mathcal{T} \leftarrow$ inverse $\mathcal{T}$
\STATE $\mathcal{I'} \leftarrow$ composite $\mathcal{T}$ over $\mathcal{I}$
\RETURN $\mathcal{I'}$\\
\centering \textbf{ALGORITHM 1}
\end{algorithmic}
%\end{algorithm}\caption{Edge Enhancement}
\vspace{0.5cm}
%
%\\
%\indent
%\begin{algorithm}
%\caption{Fancy PCA}
\begin{algorithmic}
\small
\REQUIRE Source image: $\mathcal{I}$
\STATE $\mathcal{M} \leftarrow$ Create a $255^2\times3$ matrix where the columns represent the RGB channels and all entries are the RGB values of $\mathcal{I}$.
\STATE PCA is performed on $\mathcal{M}$ using Singular Value Decomposition.
\FORALL{Pixels $I(x,y)$ in $I$}
\STATE $[\mathcal{I}_{xy}^R, \mathcal{I}_{xy}^G, \mathcal{I}_{xy}^B]^T \leftarrow$ Add $\displaystyle \frac{1}{s_p}\mathcal{P}[\alpha_1\lambda_1,\alpha_2\lambda_2,\alpha_3\lambda_3]^T$
\STATE $\bullet$ $\mathcal{P}$ is a $3\times3$ matrix where the columns are the eigenvectors
\STATE $\bullet$ $\lambda_i$ is the i\textsuperscript{th} eigenvalue corresponding to the eigenvector $[p_{i,1}, p_{i,1}, p_{i,1}]^T$
\STATE $\bullet$ $\alpha_i$ is a random variable which is drawn from a Gaussian with $0$ mean and standard deviation $0.1$
\STATE $\bullet$ $s_p$ is the scaling parameter which was initialised to $5 \cdot 10^6$ by trial and error.
\ENDFOR
\RETURN $\mathcal{I}$\\
\centering \textbf{ALGORITHM 2}
\end{algorithmic}
%\end{algorithm}
\vspace{0.5cm}

Fancy PCA is a scheme that performs PCA on sets of $RGB$ pixels throughout the training set by adding multiples of principles components to the training images (Algorithm 2).
In related work \cite{AlexNet} it is unclear as to whether the authors performed PCA on individual images or on the entire training set.
However, due to computation and memory constraints we adopted the former approach.

\begin{figure*}[t]
    \centering
    \includegraphics[width=0.95\textwidth]{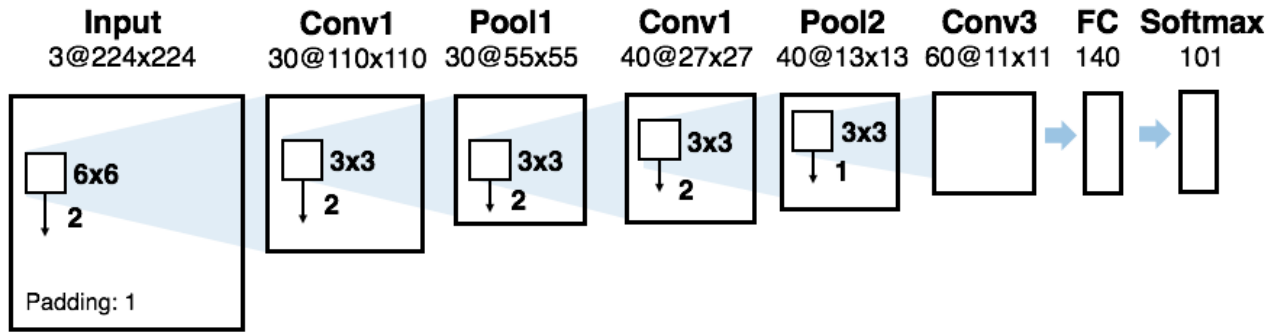}
    \caption{The \textit{Convolutional Neural Network} (CNN) architecture.}
    \label{fig:architecture}
\end{figure*}

\section{CNN Architecture}

%All the DA methods described in Section 3 were benchmarked on the CNN described in this section.
A CNN architecture was developed with the objective of obtaining a favorable tradeoff between task performance and training speed.
Training speed was crucial as the CNN had to be trained seven times on data-sets ranging from $\sim8.5$ to $\sim42.5$ thousand images using $4-$fold cross validation.  Also, the CNN had to be \textit{deep} enough (containing enough parameters) so as the network would fit the training data.\\
\indent
We used an architecture (figure \ref{fig:architecture}) similar to that described by \cite{VisualizingCNN}, where exploratory experiments indicated
a reasonable tradeoff between topology size and training speed.
The architecture consisted of $5$ trainable layers: $3$ convolutional layers, $1$ fully connected layer and $1$ softmax layer.
The CNN took a $3$ channel (representing the RGB channels) $255\times225$ image as input, $30$ filters of size $6\times6$ with a stride of $2$ and a padding of $1$ were convolved over the image producing a feature map of size $30\times110\times110$.
This layer was compressed by a max-pooling filter of size $3\times3$ with stride of $2$ which reduced it to a new feature map of dimensions $30\times55\times55$. A set of $40$ filters followed with the same size and stride as before which were convolved over the layer producing a new feature map of size $40\times27\times27$.
Again a max pooling function of the same size and stride was applied which further reduced the feature map to $40\times13\times13$. This was fed into a last CNN layer using a set of $60$ filters of size $3\times3$ of stride $1$ producing a layer of $60\times11\times11$ parameters.
Finally this was fed into a fully connected layer of size $140$ which connected to a soft-max layer of size $101$.\\
\indent
Overlapping pooling was deployed which increased CNN performance by reducing over-fitting \cite{AlexNet}.
This yielded sums of overlapping neighboring groups of neurons in the same feature map.
A fully connected layer of $140$ neurons was chosen as increased sizes did not generate greater improvements in performance \cite{ReturnDevil}.
Exploratory experiments indicated that smaller layer sizes resulted in richer encodings of the distinct classes yielding better generalization.
The depth sizes of individual convolutional layers were determined by trial and error.
Further convolutional layers did not increase performance considerably and were thus omitted to reduce computational time.
It was also found that a second fully connected layer did not improve task performance.\\
\indent
All neurons used a \textit{Rectified Linear Unit} \cite{HahnloserSeung2000}
with the weights initially being initialised from a Gaussian distribution with a $0$ mean and a standard deviation of $0.01$.
An initialisation scheme known as Xavier \cite{Xavier} was deployed which mitigated slow convergence to local minima.
All weights were updated using back-propagation and stochastic gradient descent with a learning rate of $0.01$.
A variety of update functions were tested including \textit{Adam} \cite{Adam} and \textit{Adagard} \cite{Adagrad}, however
we selected \textit{Nesterov} \cite{Nesterov} with a momentum of $0.90$ which we found to converge relatively fast and not suffer
from from numerical instability and stagnating convergence.
Additionally regularization was deployed in the form of gradient normalization with a \textit{L2} of $5 \cdot 10^{-4}$ to reduce over-fitting.
Hyper-parameters for \textit{activation function}, \textit{learning rate}, \textit{optimisation algorithm} and \textit{updater}
were based on those used in related work \cite{AlexNet} and all other parameter values were determined by exploratory experiments.
All CNN parameters used in this study are presented in table \ref{table:hyper}.
%which enabled faster convergence over other methods yet suffered from numerical instability, and
%which also showed impressive convergence but quickly stagnated without further improvements.
%The final model made use of

\begin{table}[t]
\centering
 \begin{tabular}{ c c c }
 \hline
 \textbf{Hyper-parameter}              & \textbf{Type} \\ [0.5ex]
 \hline\hline
 Activation Function                   & ReLu \\
 Weight Initialisation                 & Xavier \\
 Learning Rate                         & 0.01 \\
 Optimisation Algorithm                & SGD \\
 Updater                               & Nesterov \\
 Regularization                        & L2 Gradient Normalization \\
 Minibatch                             & 16 \\
 Epochs                                & 30 \\[0.5ex]
 \hline
 \end{tabular}
 \caption{Hyper-parameters used by the CNN in this study.}\label{table:hyper}
\end{table}

\newpage

\section{Experiments}\label{sec:exps}

This study evaluated various DA methods on the Caltech101 data-set, which is a coarse-grained data-set consisting of
$102$ categories containing a total of $9144$ images.
The Caltech101 data-set was chosen as it is a well established data-set for training CNNs containing
a large amount of varying classes \cite{ReturnDevil,VisualizingCNN}.
For CNN training most images in the Caltech101 data-set were used.  That is, $725$ images, including the \textit{background}
category in the Caltech101 data-set as it contained many uncorrelated images, were omitted.
Also, further trimming was applied
such that the cardinality of every class was divisible by $4$ for cross-validation, which further increased the number
images excluded, meaning that $8421$ images in total were evaluated.
We elected to use cross-validation which maximized
the use of selected images and better estimated how the CNNs performance would scale to other unknown data-sets.
Specifically 4-fold\footnote{Higher fold cross validation would have taken too long to train.}
cross-validation was used which partitioned the data-set into $4$ equal sized subsets, where $3$ subsets were used for training and the
other for validation purposes.\\
%As data-set classes ranged in sizes from $31$ to $800$ all were trimmed such that the cardinality was divisible by $4$
%(as 4-fold cross-validation required equally sized subsets).\\
\indent
All images within the data-set were transformed to a size of $256\times256$,
where every image was downsized such that the largest dimension was equal to $256$.
This downsized image was then centrally drawn on top of a $256\times256$ black
image\footnote{Numeric value of 0 for all channels thus acting as zero padding.}.
This enabled all augmentation schemes to have access to the full image in a fixed resolution of $256\times256$.
Finally the transformed images underwent normalization by scaling all pixel values from $[0, 255] \rightarrow [0, 1]$.\\
\indent
Every CNN was trained using $30$ epochs.
This value was determined by exploratory experiments that evaluated validation and test scores every epoch.
All implementation was completed in \textit{Java 8} using \textit{DL4j}\footnote{$deeplearning4j.org$}
with the experiments being conducted
on a \textit{NVIDIA Tesla K80} GPU using \textit{CUDA}.
All source code and experiment details can be found online\footnote{$github.com/webstorms/AugmentedDatasets$}.

\section{Results and Discussion}

%Training the CNNs on a GPU enabled the experiments to completed within 4 days (which is an order of magnitude faster compared to training on a $2.3$ GHz Intel Core i5).

\begin{table}[t]
\centering
 \begin{tabular}{ c c c }
 \hline
 & \textbf{Top-1 Accuracy} & \textbf{Top-5 Accuracy} \\ [0.5ex]
 \hline\hline
 Baseline & 48.13 $\pm$ 0.42\% & 64.50 $\pm$ 0.65\%\\
 Flipping & 49.73 $\pm$ 1.13\% & 67.36 $\pm$ 1.38\%\\
 Rotating & 50.80 $\pm$ 0.63\% & 69.41 $\pm$ 0.48\%\\
 Cropping & \textbf{61.95 $\pm$ 1.01}\% & \textbf{79.10 $\pm$ 0.80\%}\\
 Color Jittering & \textbf{49.57 $\pm$ 0.53\%} & 67.18 $\pm$ 0.42\%\\
 Edge Enhancement & 49.29 $\pm$ 1.16\% & 66.49 $\pm$ 0.84\%\\
 Fancy PCA & 49.41 $\pm$ 0.84\% & \textbf{67.54 $\pm$ 1.01\%}\\[0.5ex]
 \hline
 \end{tabular}
 \caption{CNN training results for each DA method.}\label{table:Results}
\end{table}

Table \ref{table:Results} presents experimental results, where Top-1 and Top-5 scores were
evaluated as percentages as done in the \textit{Imagenet} competition \cite{ImageNet}, though
we report accuracies rather than error rates.  The CNN's output is a multinomial distribution over
all classes:
\[ \sum p_{class} = 1 \]

The Top-1 score is the number of times the highest probability is associated with the correct target over all testing images.
The Top-5 score is the number of times the correct label is contained within the $5$ highest probabilities.
As the CNN's accuracy was evaluated using cross-validation a standard deviation was associated with every result, thus indicating
how variable the result is over different testing folds.
In table \ref{table:Results}, Top-1 and Top-5 scores in the geometric and photometric DA category are represented in bold.\\
\indent
Results indicate that in all cases of applying DA, CNN classification task performance increased.
Notably, the \textit{geometric augmentation} schemes outperformed the \textit{photometric schemes}
in terms of Top-1 and Top-5 scores.
The exception was the \textit{flipping} scheme Top-5 score being
inferior to the \textit{fancy PCA} Top-5 score.
For all schemes, a standard deviation of $0.5\% \sim 1\%$ indicated similar results over all folds with
the cross-validation.\\
\indent
The cropping scheme yielded the greatest improvement in Top-1 score with an improvement of $13.82\%$ in classification accuracy.
Results also indicated that Top-5 classification yielded a similar task improvement which corroborated related
work \cite{ReturnDevil,AircraftDA}.
We theorize that the \textit{cropping} scheme outperforms the other methods as it generates more sample images than the other augmentation schemes.
This increase in training data reduces the likelihood of over-fitting, improving generalization and thus increasing overall classification
task performance.
Also, cropping represents specific translations allowing the CNN exposure to a greater receptive view of training images
which the other augmentation schemes do not take advantage of \cite{ReturnDevil}.
\indent
%However, the \textit{rotating} and \textit{flipping} methods, although satisfactory,
%did not see as good improvements as the \textit{cropping} scheme as they are better suited for fine-grained data-sets as described in section 2.\\
%However the returns gained from sampling multiple crops are diminishing as reported by \cite{ReturnDevil}.

However, the \textit{photometric} augmentation methods yielded modest improvements in performance compared to the geometric schemes,
indicating the CNN yields increased task performance when trained on images containing invariance in geometry rather than lighting and color.
The most appropriate photometric schemes were found to be \textit{color jittering} with a top-1 classification improvement of $1.44\%$
and \textit{fancy PCA} which improved top-5 classification by $3.04\%$.
\textit{Fancy PCA} increased top-1 performance by $1.28\%$ which supported the findings of previous work \cite{AlexNet}.

We also hypothesize that \textit{color jittering} outperformed the other photometric schemes in top-1 classification as this scheme
generated augmented images containing more variation compared to the other methods (figure \ref{fig:overview}).
Also, \textit{edge enhancement} augmentation did not yield comparable task performance, likely due to
the overlay of the transformed image onto the source image (as described in section \ref{sec:photoMethods})
did not enhance the contours enough but rather lightened the entire image.
However, the exact mechanisms responsible for the variability of CNN classification task performance given
\textit{geometric} versus \textit{photometric} augmentation methods for coarse-grained data-sets remains the topic of ongoing research.

\section{Conclusion}
%% Has not been edited since V1
This study's results demonstrate that an effective method of increasing CNN classification task performance is to make use of
\textit{Data Augmentation} (DA).
Specifically, having evaluated a range of DA schemes using a relatively simple CNN architecture
we were able to demonstrate that \textit{geometric augmentation} methods outperform photometric methods
when training on a coarse-grained data-set (that is, the Caltech101 data-set).
The greatest task performance improvement was yielded by specific translations generated by the \textit{cropping} method
with a Top-1 score increase of $13.82\%$.
These results indicate the importance of augmenting coarse-grained training data-sets using transformations that alter the geometry
of the images rather than just lighting and color.\\
\indent
Future work will experiment with different coarse-grained data-sets to establish whether results obtained using the Caltech101 are transferable to other data-sets.
Additionally different CNN architectures as well as other DA methods and combinations of DA methods will be investigated in comprehensive studies
to ascertain the impact of applying a broad range of generic DA methods on coarse-grained data-sets.
%% The file named.bst is a bibliography style file for BibTeX 0.99c
\bibliographystyle{named}
\bibliography{ijcai17}

\end{document}